\documentclass{bmvc2k}
\usepackage{subcaption}
\usepackage{amsmath}
\usepackage{amssymb}
\usepackage{enumitem}
\usepackage{graphicx}
\usepackage{pdfpages}
\usepackage{booktabs}
\graphicspath{{images/uncertainty/}}
\usepackage{tabularx}
\usepackage{siunitx}
\usepackage{multirow}
\usepackage{makecell}
\usepackage{subfigure}

\usepackage{graphicx}
\usepackage{booktabs}
\usepackage{array}
\newcolumntype{P}[1]{>{\raggedright\arraybackslash}p{#1}}
\usepackage[normalem]{ulem}

\title{TPA: Temporal Prompt Alignment for Fetal
Congenital Heart Defect Classification}

\runninghead{Taratynova ET. AL}{TPA}

\addauthor{Darya Taratynova}{darya.taratynova@mbzuai.ac.ae}{1}
\addauthor{Alya Almsouti}{alya.almsouti@mbzuai.ac.ae}{2}
\addauthor{Beknur Kalmakhanbet}{beknur.kalmakhanbet@mbzuai.ac.ae}{2}
\addauthor{Numan Saeed}{numan.saeed@mbzuai.ac.ae}{2}
\addauthor{Mohammad Yaqub}{mohammad.yaqub@mbzuai.ac.ae}{2}

\addinstitution{
 Department of Machine Learning\\
 Mohamed bin Zayed University of Artificial Intelligence (MBZUAI)\\
 Abu Dhabi, UAE
}
\addinstitution{
 Department of Computer Vision\\
 Mohamed bin Zayed University of Artificial Intelligence (MBZUAI)\\
 Abu Dhabi, UAE
}

\begin{document}

\maketitle

\begin{abstract}
Congenital heart defect (CHD) detection in ultrasound videos is hindered by image noise and probe positioning variability. While automated methods can reduce operator dependence, current machine learning approaches often neglect temporal information, limit themselves to binary classification, and do not account for prediction calibration. We propose Temporal Prompt Alignment (TPA), a method leveraging foundation image-text model and prompt-aware contrastive learning to classify fetal CHD on cardiac ultrasound videos. TPA extracts features from each frame of video subclips using an image encoder, aggregates them with a trainable temporal extractor to capture heart motion, and aligns the video representation with class-specific text prompts via a margin-hinge contrastive loss. To enhance calibration for clinical reliability, we introduce a Conditional Variational Autoencoder Style Modulation (CVAESM) module, which learns a latent style vector to modulate embeddings and quantifies classification uncertainty. Evaluated on a private dataset for CHD detection and on a large public dataset, EchoNet-Dynamic, for systolic dysfunction, TPA achieves state-of-the-art macro F1 scores of 85.40\% for CHD diagnosis, while also reducing expected calibration error by 5.38\% and adaptive ECE by 6.8\%. On EchoNet-Dynamic’s three-class task, it boosts macro F1 by 4.73\% (from 53.89\% to 58.62\%). Temporal Prompt Alignment (TPA) is a framework for fetal congenital heart defect (CHD) classification in ultrasound videos that integrates temporal modeling, prompt-aware contrastive learning, and uncertainty quantification. The code is available at \url{https://github.com/BioMedIA-MBZUAI/TPA}.
\end{abstract}

\section{Introduction}
\label{sec:intro}
\begin{figure}[t]
  \centering

  \begin{minipage}[t]{0.30\textwidth}
    \centering
    \includegraphics[width=\linewidth]{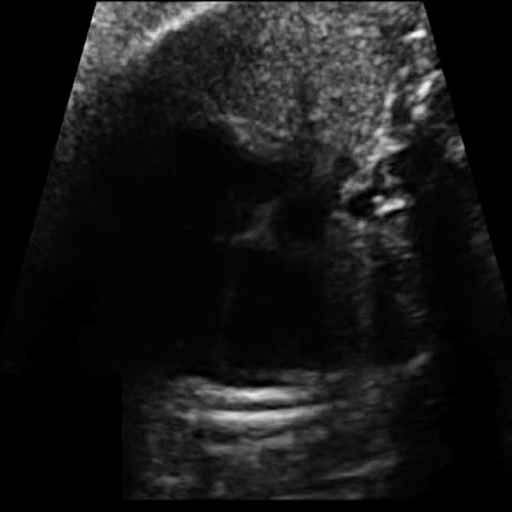}\\
    \textbf{(a)} Acoustic shadowing
    \label{fig:challenge_acoustic}
  \end{minipage}
  \hfill
  \begin{minipage}[t]{0.30\textwidth}
    \centering
    \includegraphics[width=\linewidth]{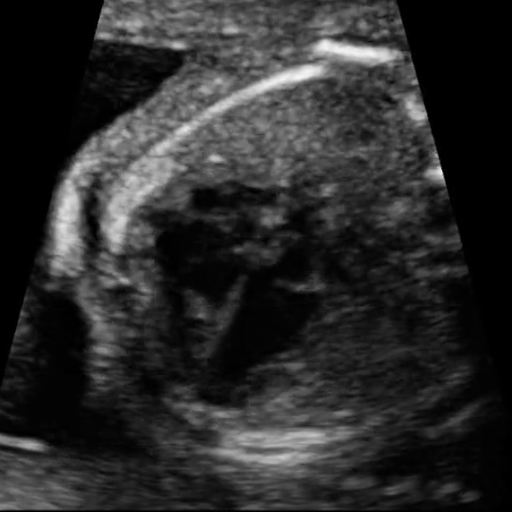}\\
    \textbf{(b)} Speckle noise
    \label{fig:challenge_speckle}
  \end{minipage}
  \hfill
  \begin{minipage}[t]{0.30\textwidth}
    \centering
    \includegraphics[width=\linewidth]{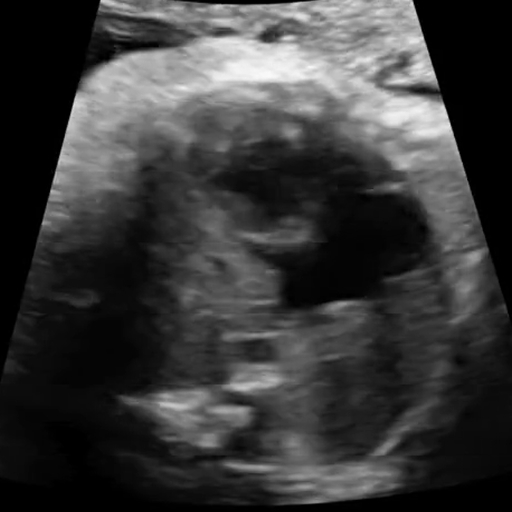}\\
    \textbf{(c)} Motion
    \label{fig:challenge_blur}
  \end{minipage}

  \vspace{3mm}
\caption{Fetal four‐chamber ultrasound frames showing common imaging artifacts.}
  \label{fig:ultrasound_challenges}
\end{figure}

Congenital heart defect (CHD) is the most common fetal anomaly, affecting approximately 1.0\% of live births, with half of those cases requiring surgical intervention \cite{vanderlinde2011birth,carvalho2002improving}. In such cases, early and precise screening is vital for improving outcomes \cite{khalil2013fetal,Liu2019}. Fetal ultrasound remains the standard screening tool in clinical practice, offering accessible cardiac imaging \cite{Yun2011}. However, its effectiveness relies heavily on the sonographer’s skill: even slight variations in probe angle or applied pressure can obscure key anatomical structures, potentially leading to missed or delayed diagnoses \cite{SHARMA2021101973}. As illustrated in Figure~\ref{fig:ultrasound_challenges}, ultrasound suffers from common artifacts, acoustic shadowing that hides parts of the heart, speckle noise that obscures anatomical boundaries, and motion that blurs moving structures, making it challenging to locate and measure chambers consistently \cite{perperidis2017dynamic}. These operator-dependent limitations underscore the need for automated CHD screening tools that can provide consistent and reliable analysis to assist clinicians.

Previous efforts to assist in cardiac ultrasound analysis have used machine-learning methods for view-plane detection \cite{an2021category,qiao2022flds,ronneberger2015u} and heart-chamber segmentation \cite{tenfetalseg,patra2020hierarchical,lu2024yolox,opticalflow,hybridclass}, primarily focusing on standard anatomical assessments. Shifting focus to CHD detection, \cite{Arnaout2021} made the first attempt by classifying individual frames from multiple cardiac planes using an ensemble of neural networks. However, many defects become apparent only through cardiac motion changes \cite{meng2024congenital}. For example, conditions such as transposition of the great arteries or hypoplastic left heart syndrome exhibit abnormal spatial or temporal motion patterns that are difficult to capture from static frames alone. Acknowledging this, \cite{saha2025selfsupervisednormalitylearningdivergence} learns healthy heart dynamics from video sequences but is limited to binary classification. Thus, extending detection to multiple defect types while modeling temporal motion remains an open challenge. 

Recently, vision–language foundation models have begun transforming ultrasound interpretation. For example, EchoCLIP \cite{christensen2024vision} delivers competitive zero‐shot performance on adult echocardiography, including implanted‐device classification, pathology detection, functional measures regression, and image–report retrieval. FetalCLIP \cite{fetalCLIP}, meanwhile, has shown itself to be a powerful feature extractor for downstream fetal ultrasound tasks. While these models are promising, recent work has shown that guiding them with aligned text prompts significantly enhances robustness and generalization \cite{wu2024semanticalignmentmultimodallarge}. In the context of CHD detection, vision–language models offer the unique ability to incorporate task-specific clinical knowledge via prompts, helping models reason more like clinicians when interpreting complex anatomical structures. Such prompts help the model focus on relevant features, and moreover, class-specific prompts further support fine-grained and few-shot adaptation \cite{long2023taskorientedmultimodalmutualleaning,ge2023improvingzeroshotgeneralizationrobustness,khattak2023maplemultimodalpromptlearning}. 

Nevertheless, accurate CHD screening requires modeling the temporal dynamics of heart motion. Early work using recurrent networks such as LSTMs laid the foundation for sequence learning \cite{lstm}, while xLSTMs extended this approach with exponential gating to capture long-range dependencies more effectively \cite{beck2024xlstmextendedlongshortterm}. Attention-based architectures have emerged to directly extract spatiotemporal features from video streams \cite{guo2024mmsummary,jaegle2021perceivergeneralperceptioniterative}, and graph-based methods further enhance this by linking frames to preserve both local motion patterns and global context \cite{chaves2024videosagevideosummarizationgraph}. Integrating these rich, pre-trained visual representations with strong temporal modeling enables the motion-aware CHD detectors.

Yet in clinical settings, reliable confidence estimates are as crucial as accurate disease detection. In medical imaging, uncertainty estimation, through Bayesian neural networks, deep ensembles, and test-time augmentation, has predominantly focused on segmentation tasks \cite{Landgraf_2024,lei2024uncertaintymodelingultrasoundimage,abutalip2024edueexpertdisagreementguidedonepass,huang2018efficientuncertaintyestimationsemantic}. In general computer vision, \cite{ullah2024cvae_sm} introduced Conditional Variational Style Modulation for image segmentation by injecting a learned style code directly into feature maps. Building on these approaches, recent efforts have extended foundation models with probabilistic adapters and contrastive uncertainty modules to enable robust uncertainty quantification across a broader set of vision tasks \cite{landgraf2025criticalsynthesisuncertaintyquantification,upadhyay2023probvlmprobabilisticadapterfrozen,zhang2022contrastiveadaptersfoundationmodel}. However, none of these methods incorporate uncertainty estimation into video-level classification.

To address these gaps, we present a unified framework that combines prompt-guided spatiotemporal modeling with uncertainty estimation for fetal ultrasound video classification.

The primary contributions are:
\begin{itemize}
    \item We propose a novel classification framework that combines vision–language foundation models with trainable temporal modeling. By integrating a lightweight temporal extractor and a prompt-aware contrastive loss, our method effectively captures spatiotemporal dynamics and aligns video content with text prompts.  
    \item We propose the first Conditional Variational Autoencoder with Style Modulation for video classification, operating on temporal video embeddings to enable reliable video-level uncertainty estimation.
\end{itemize}

\section{Proposed Method}
\begin{figure}[t]
  \centering
  \includegraphics[width=\textwidth]{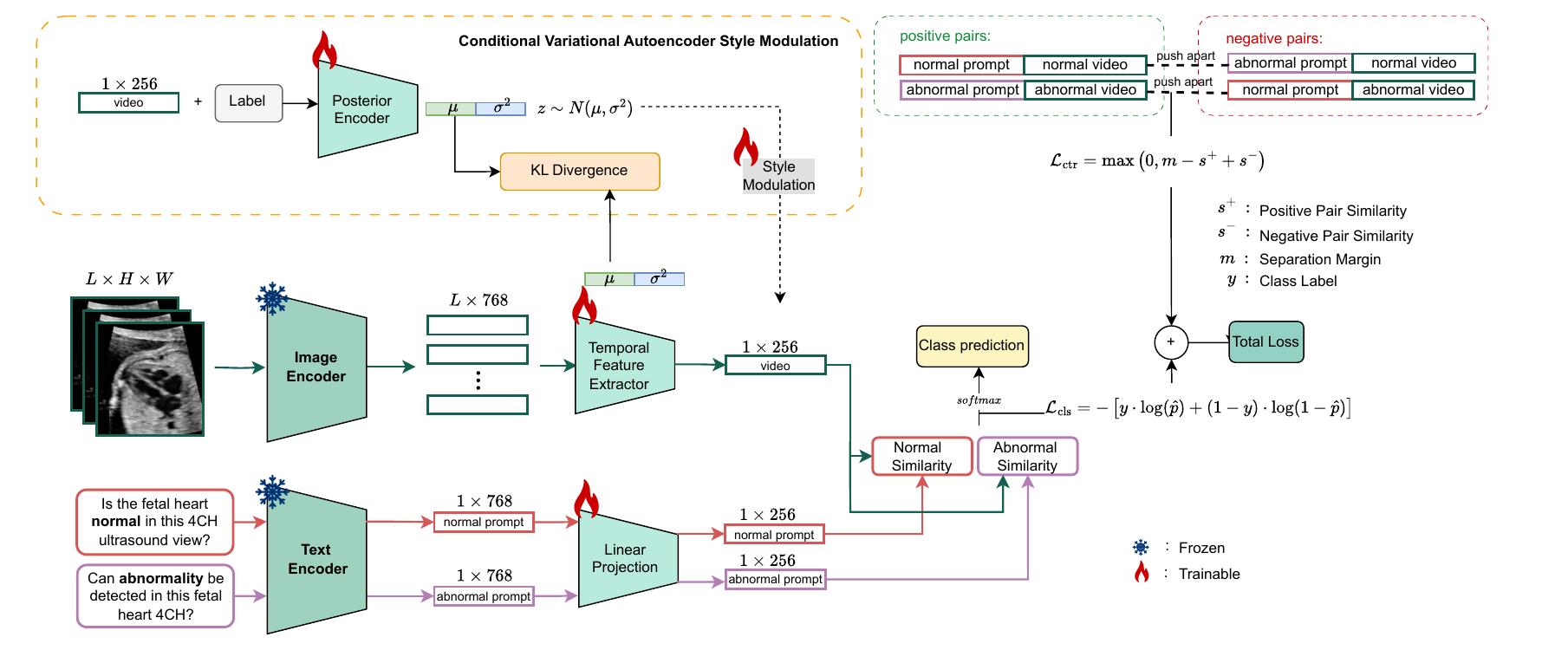}
  \vspace{1mm}
  \caption{TPA with Conditional Variational Autoencoder Style Modulation. Video frame embeddings are aggregated by a temporal extractor, matched against projected text prompt embeddings for classification and contrastive alignment. A latent style vector modulates the video features to enable uncertainty estimation.}
  \label{fig:framework}
\end{figure}

\subsection{Temporal Prompt Alignment (TPA)}

TPA leverages pretrained foundation models to encode both visual and textual inputs. In the ultrasound domain, EchoCLIP \cite{christensen2024vision} was pretrained on adult echocardiogram images paired with corresponding text reports, while FetalCLIP \cite{fetalCLIP} extends this approach to fetal ultrasound. Pretraining on aligned video–text pairs enables learning of rich spatial features and semantic associations between visual patterns and clinical concepts, which we leverage for CHD detection. TPA starts from sampling \(L\) consecutive frames from a video of \(T\) frames as shown in Figure \ref{fig:framework}. Each sampled frame is then passed through an image encoder to produce a per-frame feature vector  
\(
x_t \in \mathbb{R}^{768}.
\)
The resulting per-frame features are stacked into a matrix  
\(
  X = [\,x_1; \dots; x_L\,] \in \mathbb{R}^{L \times 768},
\)
to which a trainable temporal extractor is applied, resulting in the video embedding  
\(
  h = f_{\mathrm{temp}}(X) \in \mathbb{R}^{256}.
\)

\noindent\textbf{Binary classification.} In binary classification (\(c\in\{0,1\}\)), a text prompt is defined for each class (e.g., \textit{“Is the fetal heart normal in this 4CH ultrasound view?”} for the normal class and \textit{“Can an abnormality be detected in the fetal heart 4CH?”} for the abnormal class). Next, each prompt is passed through the text encoder to obtain  
\(
  \pi_c \in \mathbb{R}^{768},
\)
which is projected into the same space as the video embedding:  
\(
  (\pi_c)_{\mathrm{proj}} \in \mathbb{R}^{256}.
\)

Cosine similarities are then computed between the video embedding \(h\) and each projected text prompt embedding \((\pi_c)_\mathrm{proj}\), producing scores \(s_0\) and \(s_1\), which are converted to class probabilities using a temperature‐scaled softmax. The classification loss is the cross‐entropy loss, denoted \(\mathcal{L}_{\mathrm{cls}}\).

While the classification loss encourages correct predictions, it does not ensure that the video features are aligned with the corresponding prompts. To address this, we introduce a contrastive loss that explicitly separates positive and negative video-text pairs in the embedding space.  Let \(s^+\in\mathbb{R}\) denote the positive pair similarity (e.g.,  \textit{normal video and normal prompt}) and \(s^-\in\mathbb{R}\) denote the negative pair similarity (e.g., \textit{normal video and abnormal prompt}). The margin‐hinge contrastive loss, which enforces a margin \(m\) between positive and negative similarities, is defined as
\begin{equation}
\label{eq:ctr_loss}
\mathcal{L}_{\mathrm{ctr}}
=\sum_{c=0}^{1}
\max\!\bigl(0,\;m - s^+ + s^-\bigr).
\end{equation}

\noindent Finally, the total loss combines both objectives: 
\begin{equation}
\label{eq:total_loss}
\mathcal{L}_{\mathrm{total}}
=\mathcal{L}_{\mathrm{cls}}
+\alpha\,\mathcal{L}_{\mathrm{ctr}},
\end{equation}
where \(m>0\) and \(\alpha\ge0\) balance the classification and contrastive components.

\noindent\textbf{Multiclass classification.} For \(C\) classes, the positive similarity remains \(s_c^+ = s_c\), and the hardest negative is selected as 
\(\displaystyle s_c^- = \max_{j \neq c} s_j\).

\noindent The margin‐hinge contrastive loss for multiclass classification is then expressed as
\begin{equation}
\label{eq:mutliclass_ctr_loss}
  \mathcal{L}_{\mathrm{ctr}}
  = \sum_{c=0}^{C-1}
    \max\!\bigl(0,\;m - s_c^+ + s_c^-\bigr).
\end{equation}

\subsection{Conditional Variational Autoencoder Style Modulation (CVAESM)}

To enable predictive uncertainty estimation, we augment the video embedding \(h\) with a latent style representation. A latent variable \(z\in\mathbb{R}^{256}\) is introduced, with an approximate posterior conditioned on both the embedding and its class label \(y\), and a prior conditioned on the embedding alone. Both \(f_{\mathrm{post}}\) and \(f_{\mathrm{prior}}\) can be parameterized by two linear layers, enabling the learning of mean and variance functions from \(h\) and \(y\). Intuitively, the posterior captures appearance-specific variations for each video–label pair, while the prior serves to regularize the latent style space.

\noindent The posterior and prior distributions are defined as
\begin{align}
q(z\mid h,y)&=\mathcal{N}\bigl(z;\,\mu_{\mathrm{post}}(h,y),\,\sigma^2_{\mathrm{post}}(h,y)\bigr),\quad
[\mu_{\mathrm{post}},\,\log\sigma^2_{\mathrm{post}}]=f_{\mathrm{post}}([h;y]),\\
p(z\mid h)&=\mathcal{N}\bigl(z;\,\mu_{\mathrm{prior}}(h),\,\sigma^2_{\mathrm{prior}}(h)\bigr),\quad
[\mu_{\mathrm{prior}},\,\log\sigma^2_{\mathrm{prior}}]=f_{\mathrm{prior}}(h).
\end{align}

\noindent During training, samples of \(z\) are drawn using the reparameterization trick:
\[
  z = \mu_{\mathrm{post}} + \sigma_{\mathrm{post}} \odot \varepsilon,\quad
  \varepsilon\sim\mathcal{N}(0,I).
\]
At inference time, \(z\) is set to the prior mean, \(z = \mu_{\mathrm{prior}}\), for deterministic behavior.

\noindent The original video embedding is modulated by using this style code through element-wise scaling:
\begin{equation}
  \tilde{h} = h \;\odot\;\bigl(1 + g(z)\bigr),
\end{equation}
where \(g\) is a small MLP (Multilayer Perceptron). This modulation injects style-aware variation into the temporal representation, enabling better calibration of prediction uncertainty.  

The KL divergence between posterior and prior encourages regularization and disentanglement in the latent space, allowing the model to learn meaningful uncertainty signals. The final objective becomes:
\begin{equation}
\label{eq:total_loss_uncertainty}
\mathcal{L}
= \mathcal{L}_{\mathrm{cls}}
\;+\;\alpha\,\mathcal{L}_{\mathrm{ctr}}
\;+\;\beta\,D_{\mathrm{KL}}\bigl(q(z\mid h,y)\,\|\,p(z\mid h)\bigr),
\end{equation}
where \(\beta\) controls the strength of the KL term.

\section{Experimental Setup}
\noindent \textbf{Datasets and Metrics.} Experiments use a private fetal CHD dataset and EchoNet-Dynamic, evaluated by macro F1, AUC, ECE, and AECE (see Supplementary Material \ref{app:datasets}, \ref{app:eval}). The private fetal CHD dataset was acquired from a pre‐existing clinical archive, fully anonymised before access, and contained no personally identifiable information.

\noindent \textbf{Implementation.} Training details are summarized in Supplementary Material \ref{app:implementation}.

\noindent \textbf{Baselines.} We compare proposed TPA method to a range of temporal feature extractors whose outputs feed into a classification head: \emph{Frame-wise classifier}, which classifies each frame independently and aggregates via max-pooling; \emph{1D-CNN}, applying one-dimensional convolutions over frame embeddings with average and max pooling; \emph{Multi-Scale CNN}, with parallel 1D convolutions at three kernel sizes(3, 5, 7) followed by pooling and concatenation; \emph{BiLSTM} \cite{graves2005framewise}, a bidirectional LSTM; \emph{TCN} \cite{lea2016temporal}, a temporal convolutional network; \emph{xLSTM} \cite{beck2024xlstmextendedlongshortterm}, an extended LSTM; \emph{GNN}, which models frame embeddings as nodes connected within 10-frame windows using forward, backward, and undirected graphs, following the graph-based temporal structure of \cite{chaves2024videosagevideosummarizationgraph}; and \emph{CLS-token features}, where an MLP is applied to CLS tokens extracted from selected encoder layers, as described in \cite{cai2024vipllavamakinglargemultimodal}.

\section{Results and Discussion}
\subsection{Classification Performance}

\begin{table}[t]
  \centering
  \scriptsize
  \setlength{\tabcolsep}{16pt}%
  \renewcommand{\arraystretch}{1.0}

  \begin{minipage}[]{0.58\textwidth}
    \begin{tabularx}{\linewidth}{@{}l 
       S[table-format=2.2(4), separate-uncertainty]
       S[table-format=2.2(4), separate-uncertainty]@{}}
      \toprule
      \textbf{Model} & {\textbf{F1} $(\uparrow)$} & {\textbf{AUC} $(\uparrow)$} \\
      \midrule
      Frame-wise             & 73.98(2.89)  & 76.16(2.87) \\
      1D-CNN (AvgPool)       & 81.56(11.67) & 80.54(8.25) \\
      1D-CNN (MaxPool)       & 80.05(2.64)  & 80.54(8.25) \\
      Multi-Scale CNN        & 74.05(2.75)  & 65.84(7.34) \\
      BiLSTM  \cite{graves2005framewise}               & 83.44(1.25)  & 85.44(3.98) \\
      TCN   
      \cite{lea2016temporal}                 & 81.83(1.94)  & 85.44(3.98) \\
      xLSTM  \cite{beck2024xlstmextendedlongshortterm}
      & 80.32(1.99)  & 76.16(2.87) \\
      GNN   \cite{chaves2024videosagevideosummarizationgraph}
      & \underline{84.84 $\pm$ 1.64}  & \underline{87.69 $\pm$ 3.06} \\
      MultiLayer (6 + last) \cite{cai2024vipllavamakinglargemultimodal}  & 78.99(3.40)  & 78.65(5.76) \\
      MultiLayer (6,17+last) \cite{cai2024vipllavamakinglargemultimodal} & 79.37(3.50)  & 77.47(6.40) \\
      \midrule
      \textbf{TPA (ours)}    & \bfseries85.40(2.11) & \bfseries88.31(4.43) \\
      \bottomrule
    \end{tabularx}
    \vspace{3mm}
\captionof{table}{Performance comparison of baselines and TPA on binary CHD classification evaluated on the private dataset with 5-fold cross-validation. Best results are shown in \textbf{bold}.}

    \label{tab:baseline_results}
  \end{minipage}%
  \hfill%
  \begin{minipage}[]{0.38\textwidth}
    \begin{tabularx}{\linewidth}{@{}l@{\hskip 3em}
       S[table-format=1(2)] 
       S[table-format=1(2)]@{}}
      \toprule
      \textbf{Model} & {\textbf{F1} $(\uparrow)$} & {\textbf{AUC} $(\uparrow)$} \\
      \midrule
      1D-CNN        & 53.89 & \textbf{82.12} \\
      TCN \cite{lea2016temporal}            & 48.36 & 78.03 \\
      GNN \cite{chaves2024videosagevideosummarizationgraph}          & 52.45 & 78.43 \\
      xLSTM   \cite{beck2024xlstmextendedlongshortterm}      & 49.54 & 78.70 \\
      \midrule
      \textbf{TPA (ours)} & \bfseries58.62 & \bfseries82.12 \\
        \midrule
        \makecell[l]{PanEcho\\Fine-Tuned}\cite{panecho}  & 68.15 & 88.81 \\
        \midrule
      \textbf{TPA (ours)} & \bfseries68.80  & \bfseries89.16 \\
      \bottomrule
    \end{tabularx}
    \vspace{6.5mm}
\captionof{table}{Performance comparison of baselines and TPA on systolic dysfunction evaluated on the EchoNet-Dynamic test set. Best results are shown in \textbf{bold}.}
    \label{tab:echonet_results}
  \end{minipage}
\end{table}

In this section, we evaluate TPA on three classification tasks: binary CHD detection (normal vs. abnormal) on the private dataset (Table~\ref{tab:baseline_results}), multiclass systolic dysfunction classification on EchoNet-Dynamic (Table~\ref{tab:echonet_results}), and multiclass CHD classification with an increasing number of defects on the private dataset (Table~\ref{tab:all_classes_f1}). 

First, for binary CHD classification, Table~\ref{tab:baseline_results} compares TPA against baselines that first extract per-frame features with FetalCLIP \cite{fetalCLIP}, then feed these into a temporal extractor (e.g., BiLSTM, GNN) followed by a classification layer. Among these, the graph-based GNN temporal extractor achieves the highest performance, with an F1 of 84.84\% and an AUC of 87.69\%. The GNN’s graph construction, connecting nodes representing frame embeddings, effectively captures pairwise dynamics between temporally adjacent frames. Conversely, the frame-wise baseline, which does not capture the temporal dynamics, results in the lowest F1 and AUC scores. Sequence models (BiLSTM, TCN, and xLSTM) explicitly encode frame order and deliver the next-best F1 and AUC. Our TPA model shows the best result improving to an F1 of 85.40\% and an AUC of 88.31\%.


\begin{table}[ht]
  \centering
  \scriptsize
  \setlength{\tabcolsep}{20pt}%
  \sisetup{
    table-format=2.2(4),
    table-align-text-post = false,
  }
  \begin{tabularx}{\textwidth}{@{}l
    S[table-format=2.2(4)]
    S[table-format=2.2(4)]
    S[table-format=2.2(4)]
    S[table-format=2.2(4)]@{}}
    \toprule
    \textbf{Class}
      & {\textbf{TCN}} & {\textbf{GNN}}
      & {\textbf{xLSTM}} & {\textbf{TPA}} \\
    \midrule
    \multicolumn{5}{@{}l}{\bfseries Defects = 2} \\
    Macro      & 74.23(9.58)  & 72.33(6.98)  & \underline{75.42 $\pm$ 7.50}  & \bfseries78.68(8.20) \\
    Normal     & 85.62(6.99)  & 86.23(6.70)  & \underline{86.71 $\pm$ 4.99 }  & \bfseries88.69(5.24) \\
    VSD        & 44.12(1.80)  & \underline{52.27 $\pm$ 19.89} & 43.88(17.57) & \bfseries53.96(1.13) \\
    AVSD       & 61.04(19.78) & 57.19(18.32) & \bfseries61.06(19.35) & 59.56(2.06) \\
    \midrule
    \multicolumn{5}{@{}l}{\bfseries Defects = 3} \\
    Macro      & 68.14(5.40)  & \underline{70.65 $\pm$ 5.64}  & 64.79(4.51)  & \bfseries72.91(3.75) \\
    Normal     & 82.83(4.32)  & 85.10(3.00)  & \bfseries85.54(3.62)  & 84.62(2.47) \\
    VSD        & 51.10(16.96) & 50.29(9.57)  & \underline{54.34 $\pm$ 11.89} & \bfseries56.86(15.34) \\
    AVSD       & 53.38(11.42) & 53.95(11.43) & \underline{54.69 $\pm$ 9.08}  & \bfseries59.26(16.84) \\
    Arrhythmia & 39.62(12.87) & \underline{41.48 $\pm$ 9.99}  & 41.23(9.68)  & \bfseries43.99(11.12) \\
    \midrule
    \multicolumn{5}{@{}l}{\bfseries Defects = 4} \\
    Macro        & \underline{64.35 $\pm$ 4.26}  & 61.31(7.14)  & 60.60(4.95)  & \bfseries67.63(5.33) \\
    Normal       & \underline{83.38 $\pm$ 2.84}  & 82.83(2.56)  & 82.15(2.62)  & \bfseries84.72(2.77) \\
    VSD          & \underline{45.83 $\pm$ 10.41} & 41.69(9.37)  & 37.74(7.53)  & \bfseries46.70(11.88) \\
    AVSD         & \underline{49.10 $\pm$ 15.37} & 45.42(17.56) & 43.65(18.64) & \bfseries55.98(17.38) \\
    Arrhythmia   & 27.36(4.12)  & \bfseries34.43(9.37)  & 29.94(5.86)  & \underline{31.14 $\pm$ 9.88}  \\
    Cardiomegaly & \bfseries56.18(11.40) & 51.22(13.63) & 51.49(11.50) & \underline{55.68 $\pm$ 17.09} \\
  \bottomrule
  \end{tabularx}
  \vspace{3mm}
\captionof{table}{Multiclass macro F1 scores with standard deviation across 5 folds on the private dataset for 2, 3, and 4 defect classes, including the normal class. Results are shown for TCN, GNN, xLSTM baselines, and the proposed TPA method. Each setting includes the macro F1 score and class-wise performance. Best results are in \textbf{bold}; second-best macro F1 scores are \underline{underlined}.}
  \label{tab:all_classes_f1}
\end{table}

We next evaluate multiclass CHD classification as the number of congenital heart defects increases (Table~\ref{tab:all_classes_f1}). TPA achieves the highest macro‐F1 in all settings, while the second‐best shifts with number of defects (xLSTM for two, GNN for three, and TCN for four). At the class level, TPA improves Normal F1 score by approximately 2\% over the runner‐up in both the two‐ and four‐defect settings. In three‐defect case despite falling slightly behind by 0.92\%, it reduces fold‐to‐fold standard deviation by 1.15\%. For AVSD, TPA remains strongest in the three‐ and four‐defect scenarios, and although its F1 is lower in the two‐defect case, it cuts standard deviation by 17.29\%. In VSD, TPA consistently leads, outperforming the next best by an average of 1.68\% across all settings. In the two-defect task, TPA not only achieves the highest VSD F1 but also cuts its standard deviation from 19.89\% to 1.13\%. This indicates that, even when TPA’s F1 is marginally below the top score, its performance is far more consistent across folds. For Arrhythmia class, TPA improves F1 by 1.51\% over the next best method; however, in the four-defect setting, its performance on both Arrhythmia and Cardiomegaly falls just below the top baseline. This reflects the difficulty of achieving consistently high performance across all defect categories, particularly for rarer conditions with limited training samples.

\begin{figure}[ht]
  \centering
  \includegraphics[width=0.40\linewidth]{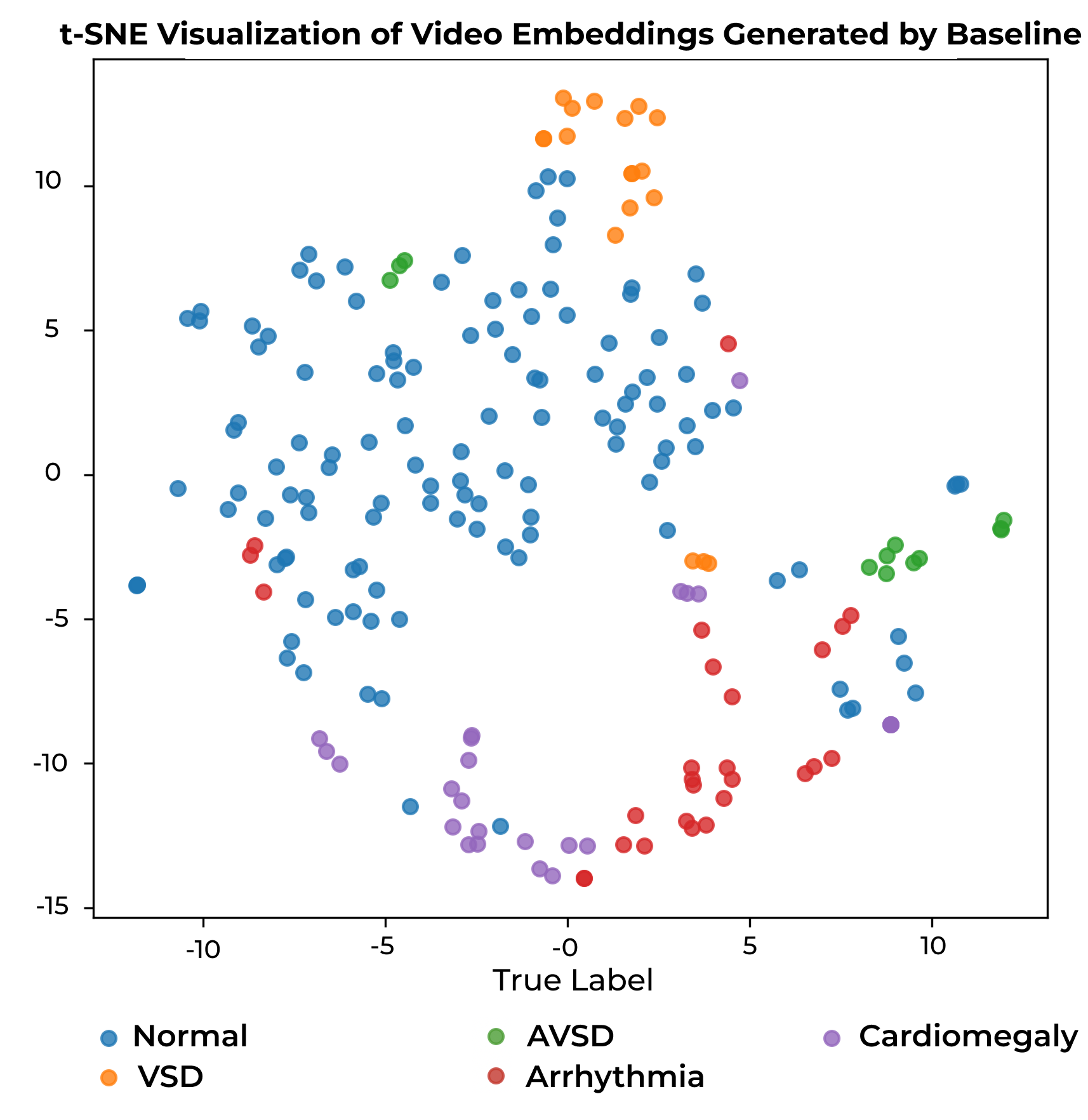}
  \includegraphics[width=0.40\linewidth]{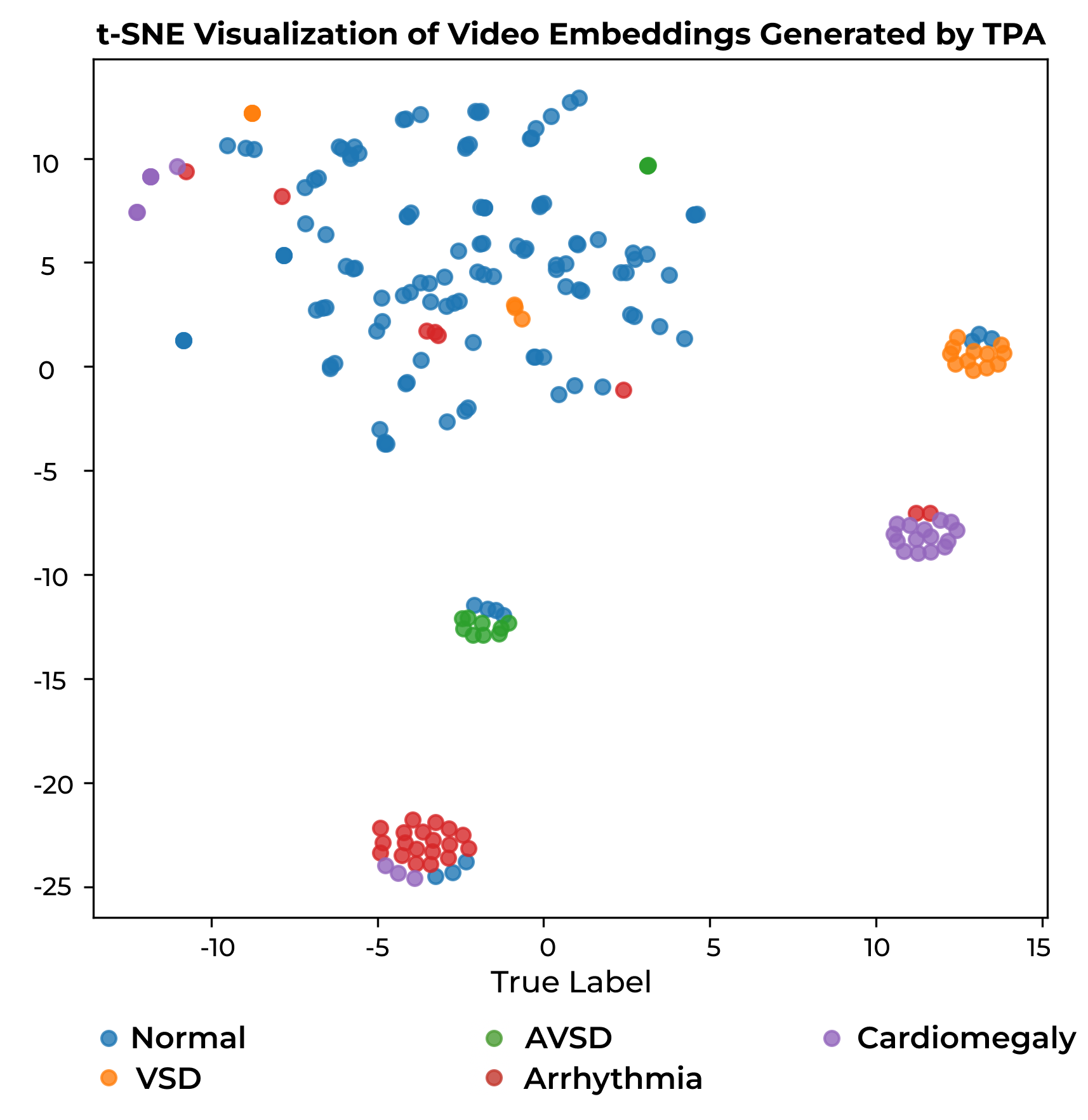}
  \vspace{2mm}
\caption{t-SNE visualizations of video embeddings from the xLSTM (left) and TPA (right).}
  \label{fig:tsne_visualizations}
\end{figure}

In addition to the quantitative results, Figure~\ref{fig:tsne_visualizations} visualizes the video embeddings for the three-defects task using t-SNE, with each point colored by its ground-truth label. By aligning each video representation with class‐specific text prompts, TPA leverages semantic guidance to pull samples toward their textual prototypes. As a result, most samples form well-separated clusters with only a few overlaps. Compared to the baseline, TPA results in noticeably more distinct and semantically coherent groupings.

We also validate our method on the EchoNet-Dynamic dataset for systolic dysfunction, given the absence of a publicly available CHD benchmark. The results in Table \ref{tab:echonet_results} are split into two sections. In the upper section, TPA uses EchoCLIP \cite{christensen2024vision} to extract image and text embeddings, followed by the temporal extractor, outperforming the best baseline (1D-CNN), improving F1 from 53.89\% to 58.62\%. In the lower section, we replace EchoCLIP\cite{christensen2024vision} with PanEcho \cite{panecho}, a video feature extractor for adult echocardiography, and use ClipMD \cite{glassberg2023increasingtextualcontextsize} as the text encoder for TPA, comparing against a fine-tuned PanEcho model. In this setting, TPA also raises F1 from 68.15\% to 68.80\% and AUC from 88.81\% to 89.16\%.

\begin{figure}[t]
  \centering
  \begin{minipage}{\textwidth}
    \centering
    \includegraphics[width=0.40\linewidth]{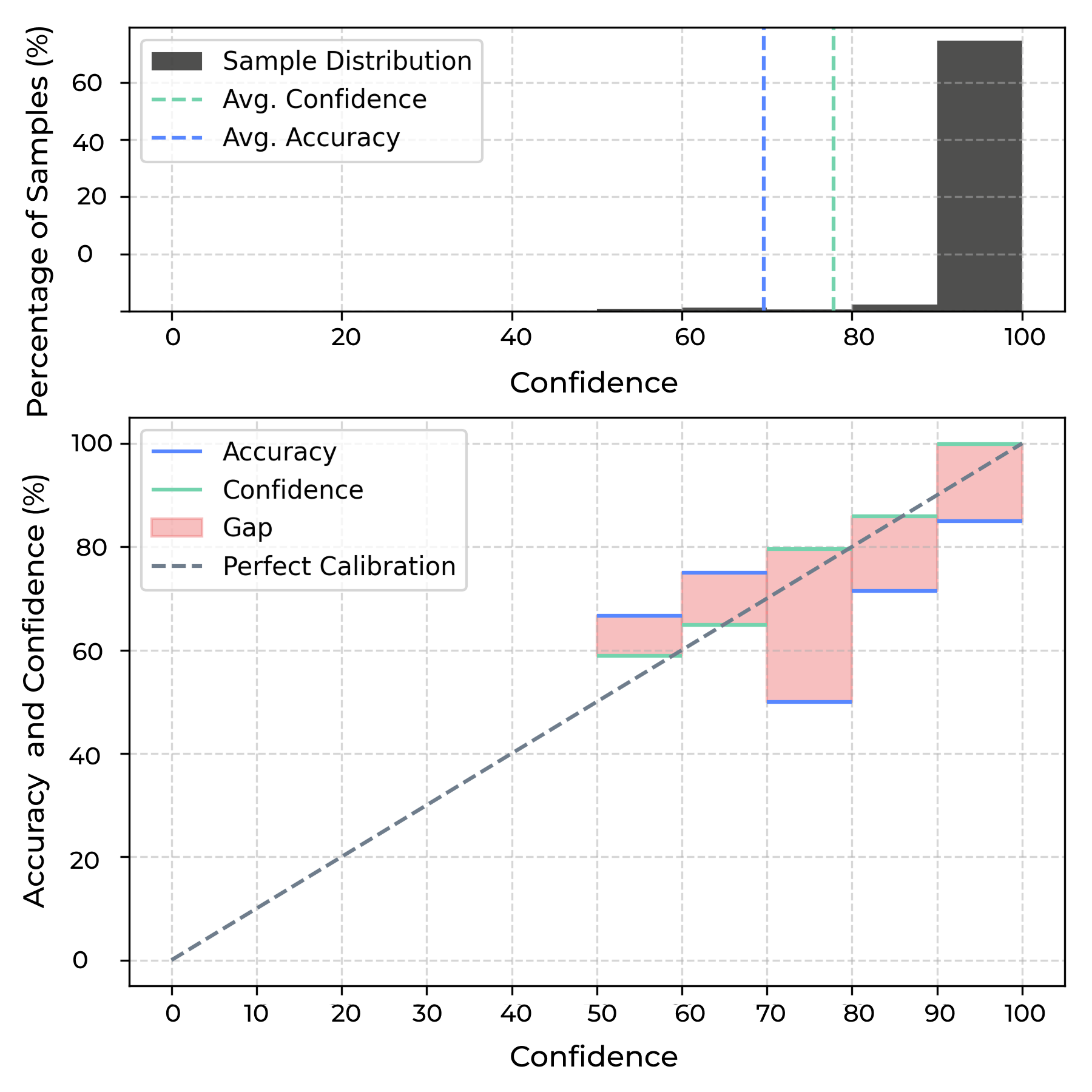}
    \includegraphics[width=0.40\linewidth]{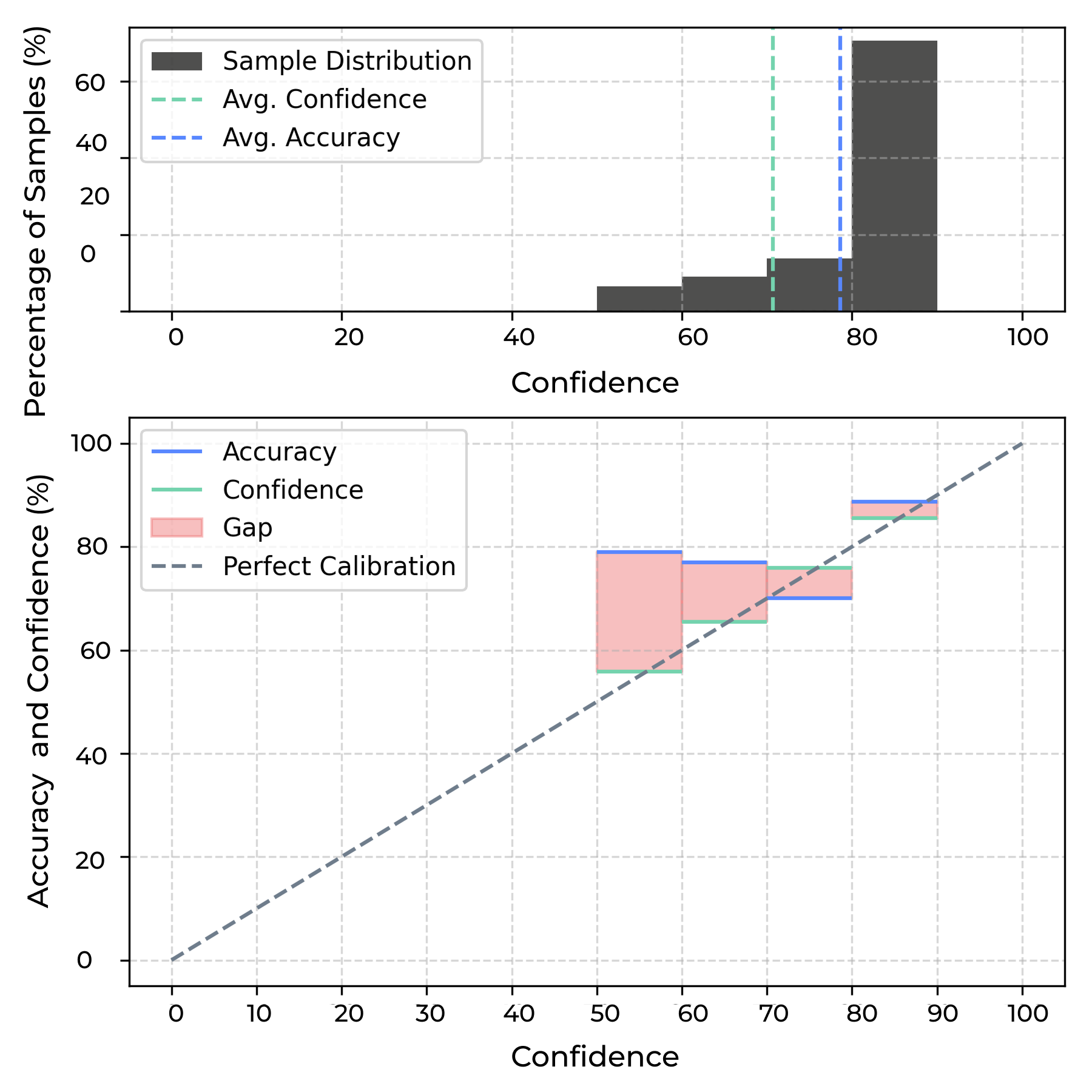}
    \vspace{2mm}

   \caption{Calibration curves before (left) and after (right) applying CVAESM.}
    \label{fig:calibration_curves}
  \end{minipage}
      \vspace{2mm}

  \begin{minipage}{\textwidth}
    \centering
    \scriptsize
    \renewcommand{\arraystretch}{1.4}
    \begin{tabular}{@{}lccc@{}}
      \toprule
      \textbf{Method}    & \textbf{F1} ($\uparrow$)       & \textbf{AECE} ($\downarrow$)     & \textbf{ECE} ($\downarrow$)     \\
      \midrule
      Before             & $\mathbf{85.40 \pm 2.36}$      & $39.76 \pm 4.81$                  & $15.98 \pm 3.83$                \\
      After CVAESM       & $84.12 \pm 1.52$               & $\mathbf{32.91 \pm 4.49}$         & $\mathbf{10.60 \pm 4.40}$       \\
      \bottomrule
    \end{tabular}
    \vspace{2mm}

    \captionof{table}{Quantitative evaluation of TPA’s classification accuracy and calibration performance across 5 folds, before and after integrating the CVAESM module. Best result is in \textbf{bold}.}

    \label{tab:cal_and_metrics}
  \end{minipage}
\end{figure}

\subsection{Calibration Performance}
TPA is equipped with a lightweight uncertainty quantification module (CVAESM), and we evaluate its effect on model calibration both quantitatively and qualitatively. Table \ref{tab:cal_and_metrics} shows that adding CVAESM reduces AECE by approximately 7\% and ECE by 5\%. Although the mean F1 score decreases slightly, its standard deviation across folds also drops, indicating more consistent performance. Qualitatively, Figure~\ref{fig:calibration_curves} illustrates how CVAESM reshapes both the distribution of confidence scores and the model’s reliability. Before adding CVAESM, more than 70\% of predictions cluster in the 90–100\% confidence bin, yet their true accuracy is below. This overconfidence creates large positive gaps between confidence and accuracy, and the reliability curve lies well above the ideal diagonal. After adding CVAESM, confidence scores spread into lower bins of 60–90\%, and the reliability curve tracks the diagonal more closely. The gap bars in high-confidence regions shrink substantially, showing that reported probabilities now better align with observed accuracy.

\subsection{Ablation}

\noindent \textbf{Effect of Margin and Contrastive Loss.} Equation~\ref{eq:total_loss} introduces two key hyperparameters: the margin $m$, which specifies how far positive pairs should be pushed from negative pairs, and the contrastive loss weight $\alpha$, which balances the contrastive term against the classification loss. To quantify their impact, we performed several experiments on binary CHD detection on the private dataset using a fixed 1D‐CNN backbone within TPA. In Supplementary Material \ref{app:ablations}, Table~\ref{tab:tpa_grid_ablation_clean}(a) reports performance for two margin values, \(m\in\{0.5,1.0\}\), and three contrastive loss weights, \(\alpha\in\{0,0.5,1\}\). A smaller margin \(m=0.5\) encourages positive pairs to cluster closely without forcing negatives so far apart that the model overfits to the limited negative examples in each batch. In contrast, \(m=1.0\) can push negatives too aggressively as seen by decreased F1 score to 84.45\% from 84.58\% with \(m=0.5\). We also have observed that $\alpha=0.5$ achieves the best trade‐off, producing an F1 of 84.69\% and an AUC of 87.59\%. If \(\alpha\) is too large, the model may overemphasize prompt alignment at the expense of class separation,  if too small, it fails to leverage the semantic guidance from text prompts.

\noindent \textbf{Effect of Prompt Randomization.} We also experimented with randomizing the text prompts at each training epoch, the result is reported in Table \ref{tab:tpa_grid_ablation_clean}(a) (Supplementary Material \ref{app:ablations}). This configuration results in F1 of 84.24\% and an AUC of 86.96\%, slightly below the 84.69\% and 87.59\% achieved with fixed prompts. Although prompt variation can act as a form of data augmentation, it introduces semantic inconsistency: class prompts shift from epoch to epoch, preventing the contrastive loss from converging on stable cluster centers.

\noindent \textbf{Effect of GNN and xLSTM Hyperparameters.} To assess their impact on TPA’s performance, we tuned both the graph‐based and LSTM extractors within the TPA. For the GNN, in Table~\ref{tab:tpa_grid_ablation_clean}(b) (Supplementary Material \ref{app:ablations}), varying the number of temporal connections (e.g., linking each node to 4 vs. 16 neighbors) had almost no impact on F1 or AUC, demonstrating robustness to graph sparsity. Switching from mean to max pooling raised F1 by about 0.5\% but caused a slight AUC drop, suggesting that max pooling’s focus on the single strongest activation can miss supporting information in other frames and hurt the model’s ranking consistency. As shown in Table~\ref{tab:tpa_grid_ablation_clean}(c) (Supplementary Material \ref{app:ablations}), the sLSTM variant of the xLSTM extractor outperforms its mLSTM counterpart, increasing F1 by 0.45\% and AUC by 0.63\%. This can be attributed to simpler gating dynamics better matching our moderate‐length ultrasound sequences. We also evaluated where to insert the sLSTM block in a three‐stage pipeline: early, mid, or late fusion. Early fusion achieved the best results, while mid and late fusion fell short by up to 1\%, indicating that capturing temporal dependencies at the outset is most effective. Finally, increasing xLSTM depth from 2 to 12 blocks provided only marginal gains, implying that a shallow recurrent stack suffices to model the periodic motion in four‐chamber fetal heart videos.  

\noindent \textbf{Effect of Temporal Extractor.} We evaluate five temporal extractors: 1D‐CNN, TCN, BiLSTM, xLSTM, and GNN, within the TPA framework on both binary CHD detection and multiclass classification (Tables~\ref{tab:tpa_grid_ablation_clean}(d) and \ref{tab:tpa_f1_auc_multiclass_grouped_tpa} in Supplementary Material \ref{app:ablations}). On private dataset for binary CHD detection, GNN leads with an F1 of 84.84\% and an AUC of 87.69\%, followed by TCN (81.83\% and 85.44\%) and xLSTM (80.32\% and 84.64\%). On EchoNet-Dynamic, GNN achieves the highest performance again, with an F1 of 58.62\% and an AUC of 82.11\%. In the multiclass CHD setting, GNN consistently achieves the highest macro‐F1 and AUC across two-, three-, and four-defect settings (e.g., 76.46\% and 72.76\% for two defects). TPA consistently improves F1 and AUC across every temporal extractor, demonstrating its flexibility to diverse sequence-modeling architectures.

\section{Conclusion}
We presented Temporal Prompt Alignment (TPA) that leverages foundation image–text models, prompt‐aware contrastive learning, and lightweight uncertainty quantification to tackle CHD classification in fetal ultrasound. TPA achieves state‐of‐the‐art discrimination and improved calibration across private CHD and EchoNet‐Dynamic benchmarks. Extensive ablations demonstrate the robustness of design both to temporal‐extractor choices and to hyperparameter settings, while t‐SNE visualizations confirm semantically coherent clustering in the learned feature space. Future work will explore richer prompt designs, leverage multiple view planes, and employ adaptive fusion of text and video modalities.

\section{Acknowledgments}
We gratefully acknowledge that this work was supported by GE Healthcare and partly funded by Sandooq Al Watan, Erth Zayed Philanthropies (Grant ID PRJ-SWARD-660).

\bibliography{egbib}

\begin{thebibliography}{46}
\providecommand{\natexlab}[1]{#1}
\providecommand{\url}[1]{\texttt{#1}}
\expandafter\ifx\csname urlstyle\endcsname\relax
  \providecommand{\doi}[1]{doi: #1}\else
  \providecommand{\doi}{doi: \begingroup \urlstyle{rm}\Url}\fi

\bibitem[Abutalip et~al.(2024)Abutalip, Saeed, Sobirov, Andrearczyk,
  Depeursinge, and Yaqub]{abutalip2024edueexpertdisagreementguidedonepass}
Kudaibergen Abutalip, Numan Saeed, Ikboljon Sobirov, Vincent Andrearczyk,
  Adrien Depeursinge, and Mohammad Yaqub.
\newblock Edue: Expert disagreement-guided one-pass uncertainty estimation for
  medical image segmentation, 2024.
\newblock URL \url{https://arxiv.org/abs/2403.16594}.

\bibitem[An et~al.(2021)An, Zhu, Wang, Zhou, Zhou, Yang, Zhang, Liu, Jiao, and
  He]{an2021category}
Shan An, Haogang Zhu, Yuanshuai Wang, Fangru Zhou, Xiaoxue Zhou, Xu~Yang,
  Yingying Zhang, Xiangyu Liu, Zhicheng Jiao, and Yihua He.
\newblock A category attention instance segmentation network for four cardiac
  chambers segmentation in fetal echocardiography.
\newblock \emph{Computerized Medical Imaging and Graphics}, 93:\penalty0
  101983, October 2021.
\newblock \doi{10.1016/j.compmedimag.2021.101983}.
\newblock Epub 2021 Sep 16; PMID: 34610500.

\bibitem[Arnaout et~al.(2021)Arnaout, Curran, Zhao, Levine, Chinn, and
  Moon-Grady]{Arnaout2021}
Rima Arnaout, Lindsey Curran, Yao Zhao, Jennifer~C. Levine, Ellen Chinn, and
  Anita~J. Moon-Grady.
\newblock An ensemble of neural networks provides expert-level prenatal
  detection of complex congenital heart disease.
\newblock \emph{Nature Medicine}, 27\penalty0 (5):\penalty0 882--891, May 2021.
\newblock \doi{10.1038/s41591-021-01342-5}.
\newblock URL \url{https://doi.org/10.1038/s41591-021-01342-5}.

\bibitem[Beck et~al.(2024)Beck, Pöppel, Spanring, Auer, Prudnikova, Kopp,
  Klambauer, Brandstetter, and Hochreiter]{beck2024xlstmextendedlongshortterm}
Maximilian Beck, Korbinian Pöppel, Markus Spanring, Andreas Auer, Oleksandra
  Prudnikova, Michael Kopp, Günter Klambauer, Johannes Brandstetter, and Sepp
  Hochreiter.
\newblock xlstm: Extended long short-term memory, 2024.
\newblock URL \url{https://arxiv.org/abs/2405.04517}.

\bibitem[Cai et~al.(2024)Cai, Liu, Park, Mustikovela, Meyer, Chai, and
  Lee]{cai2024vipllavamakinglargemultimodal}
Mu~Cai, Haotian Liu, Dennis Park, Siva~Karthik Mustikovela, Gregory~P. Meyer,
  Yuning Chai, and Yong~Jae Lee.
\newblock Vip-llava: Making large multimodal models understand arbitrary visual
  prompts, 2024.
\newblock URL \url{https://arxiv.org/abs/2312.00784}.

\bibitem[Carvalho et~al.(2002)Carvalho, Mavrides, Shinebourne, Campbell, and
  Thilaganathan]{carvalho2002improving}
J.~S. Carvalho, E.~Mavrides, E.~A. Shinebourne, S.~Campbell, and
  B.~Thilaganathan.
\newblock Improving the effectiveness of routine prenatal screening for major
  congenital heart defects.
\newblock \emph{Heart}, 88\penalty0 (4):\penalty0 387--391, October 2002.
\newblock \doi{10.1136/heart.88.4.387}.

\bibitem[Chaves and Tripathi(2024)]{chaves2024videosagevideosummarizationgraph}
Jose M.~Rojas Chaves and Subarna Tripathi.
\newblock Videosage: Video summarization with graph representation learning,
  2024.
\newblock URL \url{https://arxiv.org/abs/2404.10539}.

\bibitem[Christensen et~al.(2024)Christensen, Vukadinovic, Yuan, and
  Ouyang]{christensen2024vision}
Mads Christensen, Milan Vukadinovic, Ning Yuan, and David Ouyang.
\newblock Vision--language foundation model for echocardiogram interpretation.
\newblock \emph{Nature Medicine}, 30\penalty0 (5):\penalty0 1481--1488, 2024.
\newblock \doi{10.1038/s41591-024-02959-y}.

\bibitem[Ge et~al.(2023)Ge, Ren, Gallagher, Wang, Yang, Adam, Itti,
  Lakshminarayanan, and Zhao]{ge2023improvingzeroshotgeneralizationrobustness}
Yunhao Ge, Jie Ren, Andrew Gallagher, Yuxiao Wang, Ming-Hsuan Yang, Hartwig
  Adam, Laurent Itti, Balaji Lakshminarayanan, and Jiaping Zhao.
\newblock Improving zero-shot generalization and robustness of multi-modal
  models, 2023.
\newblock URL \url{https://arxiv.org/abs/2212.01758}.

\bibitem[Glassberg and Hope(2023)]{glassberg2023increasingtextualcontextsize}
Idan Glassberg and Tom Hope.
\newblock Increasing textual context size boosts medical image-text matching,
  2023.
\newblock URL \url{https://arxiv.org/abs/2303.13340}.

\bibitem[Graves and Schmidhuber(2005)]{graves2005framewise}
Alex Graves and J{\"u}rgen Schmidhuber.
\newblock Framewise phoneme classification with bidirectional lstm and other
  neural network architectures.
\newblock \emph{Neural Networks}, 18\penalty0 (5-6):\penalty0 602--610, 2005.
\newblock \doi{10.1016/j.neunet.2005.06.042}.

\bibitem[Guo et~al.(2017)Guo, Pleiss, Sun, and Weinberger]{guo2017calibration}
Chuan Guo, Geoff Pleiss, Yu~Sun, and Kilian~Q Weinberger.
\newblock On calibration of modern neural networks.
\newblock In \emph{International conference on machine learning}, pages
  1321--1330. PMLR, 2017.

\bibitem[Guo et~al.(2024)Guo, Men, and Noble]{guo2024mmsummary}
Xiaoqing Guo, Qianhui Men, and J~Alison Noble.
\newblock Mmsummary: Multimodal summary generation for fetal ultrasound video.
\newblock In \emph{International Conference on Medical Image Computing and
  Computer-Assisted Intervention}, pages 678--688. Springer, 2024.

\bibitem[Hochreiter and Schmidhuber(1997)]{lstm}
Sepp Hochreiter and Jürgen Schmidhuber.
\newblock Long short-term memory.
\newblock \emph{Neural Computation}, 9\penalty0 (8):\penalty0 1735--1780, 11
  1997.
\newblock ISSN 0899-7667.
\newblock \doi{10.1162/neco.1997.9.8.1735}.
\newblock URL \url{https://doi.org/10.1162/neco.1997.9.8.1735}.

\bibitem[Holste et~al.(2023)Holste, Oikonomou, Mortazavi, Coppi, Faridi,
  Miller, Forrest, McNamara, Ohno-Machado, Yuan, Gupta, Ouyang, Krumholz, Wang,
  and Khera]{ehad456}
Gregory Holste, Evangelos~K Oikonomou, Bobak~J Mortazavi, Andreas Coppi,
  Kamil~F Faridi, Edward~J Miller, John~K Forrest, Robert~L McNamara, Lucila
  Ohno-Machado, Neal Yuan, Aakriti Gupta, David Ouyang, Harlan~M Krumholz,
  Zhangyang Wang, and Rohan Khera.
\newblock Severe aortic stenosis detection by deep learning applied to
  echocardiography.
\newblock \emph{European Heart Journal}, 44\penalty0 (43):\penalty0 4592--4604,
  08 2023.
\newblock ISSN 0195-668X.
\newblock \doi{10.1093/eurheartj/ehad456}.
\newblock URL \url{https://doi.org/10.1093/eurheartj/ehad456}.

\bibitem[Holste et~al.(2025)Holste, Oikonomou, Tokodi, Kov{\'a}cs, Wang, and
  Khera]{panecho}
Gregory Holste, Evangelos~K. Oikonomou, M{\'a}rton Tokodi, Attila Kov{\'a}cs,
  Zhangyang Wang, and Rohan Khera.
\newblock Panecho: Complete ai-enabled echocardiography interpretation with
  multi-task deep learning.
\newblock \emph{medRxiv}, 2025.
\newblock \doi{10.1101/2024.11.16.24317431}.
\newblock URL
  \url{https://www.medrxiv.org/content/early/2025/04/16/2024.11.16.24317431}.

\bibitem[Huang et~al.(2018)Huang, Hsu, Chiu, Wu, and
  Sun]{huang2018efficientuncertaintyestimationsemantic}
Po-Yu Huang, Wan-Ting Hsu, Chun-Yueh Chiu, Ting-Fan Wu, and Min Sun.
\newblock Efficient uncertainty estimation for semantic segmentation in videos,
  2018.
\newblock URL \url{https://arxiv.org/abs/1807.11037}.

\bibitem[Jaegle et~al.(2021)Jaegle, Gimeno, Brock, Zisserman, Vinyals, and
  Carreira]{jaegle2021perceivergeneralperceptioniterative}
Andrew Jaegle, Felix Gimeno, Andrew Brock, Andrew Zisserman, Oriol Vinyals, and
  Joao Carreira.
\newblock Perceiver: General perception with iterative attention, 2021.
\newblock URL \url{https://arxiv.org/abs/2103.03206}.

\bibitem[Khalil and Nicolaides(2013)]{khalil2013fetal}
A.~Khalil and K.~H. Nicolaides.
\newblock Fetal heart defects: potential and pitfalls of first-trimester
  detection.
\newblock \emph{Seminars in Fetal \& Neonatal Medicine}, 18\penalty0
  (5):\penalty0 251--260, October 2013.
\newblock \doi{10.1016/j.siny.2013.05.004}.
\newblock Epub 2013 Jun 7; PMID: 23751926.

\bibitem[Khattak et~al.(2023)Khattak, Rasheed, Maaz, Khan, and
  Khan]{khattak2023maplemultimodalpromptlearning}
Muhammad~Uzair Khattak, Hanoona Rasheed, Muhammad Maaz, Salman Khan, and
  Fahad~Shahbaz Khan.
\newblock Maple: Multi-modal prompt learning, 2023.
\newblock URL \url{https://arxiv.org/abs/2210.03117}.

\bibitem[Landgraf et~al.(2024)Landgraf, Wursthorn, Hillemann, and
  Ulrich]{Landgraf_2024}
Steven Landgraf, Kira Wursthorn, Markus Hillemann, and Markus Ulrich.
\newblock Dudes: Deep uncertainty distillation using ensembles for semantic
  segmentation.
\newblock \emph{PFG – Journal of Photogrammetry, Remote Sensing and
  Geoinformation Science}, 92\penalty0 (2):\penalty0 101–114, March 2024.
\newblock ISSN 2512-2819.
\newblock \doi{10.1007/s41064-024-00280-4}.
\newblock URL \url{http://dx.doi.org/10.1007/s41064-024-00280-4}.

\bibitem[Landgraf et~al.(2025)Landgraf, Qin, and
  Ulrich]{landgraf2025criticalsynthesisuncertaintyquantification}
Steven Landgraf, Rongjun Qin, and Markus Ulrich.
\newblock A critical synthesis of uncertainty quantification and foundation
  models in monocular depth estimation, 2025.
\newblock URL \url{https://arxiv.org/abs/2501.08188}.

\bibitem[Lea et~al.(2016)Lea, Vidal, Reiter, and Hager]{lea2016temporal}
Colin Lea, Rene Vidal, Austin Reiter, and Gregory~D. Hager.
\newblock Temporal convolutional networks: A unified approach to action
  segmentation, 2016.

\bibitem[Lei(2024)]{lei2024uncertaintymodelingultrasoundimage}
Shuge Lei.
\newblock Uncertainty modeling in ultrasound image segmentation for precise
  fetal biometric measurements, 2024.
\newblock URL \url{https://arxiv.org/abs/2401.09639}.

\bibitem[Liu et~al.(2019)Liu, Chen, Zühlke, Black, Choy, Li, and
  Keavney]{Liu2019}
Yingjuan Liu, Sen Chen, Liesl Zühlke, Graeme~C. Black, Mun-kit Choy, Ningxiu
  Li, and Bernard~D. Keavney.
\newblock Global birth prevalence of congenital heart defects 1970–2017:
  updated systematic review and meta-analysis of 260 studies.
\newblock \emph{International Journal of Epidemiology}, 48\penalty0
  (2):\penalty0 455--463, 2019.
\newblock \doi{10.1093/ije/dyz009}.

\bibitem[Long et~al.(2023)Long, Zhao, Yuan, Tan, Liu, Zhou, Wang, and
  Wang]{long2023taskorientedmultimodalmutualleaning}
Sifan Long, Zhen Zhao, Junkun Yuan, Zichang Tan, Jiangjiang Liu, Luping Zhou,
  Shengsheng Wang, and Jingdong Wang.
\newblock Task-oriented multi-modal mutual leaning for vision-language models,
  2023.
\newblock URL \url{https://arxiv.org/abs/2303.17169}.

\bibitem[Lu et~al.(2024)Lu, Li, Pu, Tan, and Zhu]{lu2024yolox}
Yuhuan Lu, Kenli Li, Bin Pu, Ying Tan, and Ningbo Zhu.
\newblock A yolox-based deep instance segmentation neural network for cardiac
  anatomical structures in fetal ultrasound images.
\newblock \emph{IEEE/ACM Transactions on Computational Biology and
  Bioinformatics}, 21\penalty0 (4):\penalty0 1007--1018, Jul-Aug 2024.
\newblock \doi{10.1109/TCBB.2022.3222356}.
\newblock Epub 2024 Aug 8; PMID: 36378800.

\bibitem[Maani et~al.(2025)Maani, Saleem, Alasmawi, Mohammed, and
  Valappi]{fetalCLIP}
Saeed Maani, Farooq Saleem, Diehl Alasmawi, Waring Mohammed, and Bricker
  Valappi.
\newblock Fetalclip: A visual-language foundation model for fetal ultrasound
  image analysis --- arxiv.org.
\newblock \url{https://arxiv.org/html/2502.14807v1}, 2025.

\bibitem[Meng et~al.(2024)Meng, Song, Zhang, Lu, Li, Zhang, and
  Zhang]{meng2024congenital}
Xiangyu Meng, Mengdi Song, Kai Zhang, Wenhao Lu, Yujia Li, Cheng Zhang, and
  Yuhan Zhang.
\newblock Congenital heart disease: types, pathophysiology, diagnosis, and
  treatment options.
\newblock \emph{MedComm (2020)}, 5\penalty0 (7):\penalty0 e631, Jul 2024.
\newblock \doi{10.1002/mco2.631}.

\bibitem[Nixon et~al.(2019)Nixon, Dusenberry, Zhang, Jerfel, and
  Tran]{nixon2019measuring}
Jeremy Nixon, Michael~W Dusenberry, Linchuan Zhang, Ghassen Jerfel, and Dustin
  Tran.
\newblock Measuring calibration in deep learning.
\newblock In \emph{Proceedings of the IEEE/CVF Conference on Computer Vision
  and Pattern Recognition Workshops (CVPRW)}, 2019.

\bibitem[Ouyang et~al.(2020)Ouyang, He, Ghorbani, Lungren, Colucci, Shen,
  Ashley, and Liang]{ouyang2020video}
Daniel Ouyang, Bowen He, Amirata Ghorbani, Matthew~P. Lungren, William~S.
  Colucci, Dinggang Shen, Euan~A. Ashley, and David~H. Liang.
\newblock Video-based ai for beat-to-beat assessment of cardiac function.
\newblock \emph{Nature}, 580:\penalty0 252--256, March 2020.
\newblock \doi{10.1038/s41586-020-2145-8}.
\newblock URL \url{https://doi.org/10.1038/s41586-020-2145-8}.
\newblock Received: 11 November 2019; Accepted: 20 February 2020; Published: 25
  March 2020; Issue Date: 09 April 2020.

\bibitem[Patra and Noble(2020)]{patra2020hierarchical}
Arijit Patra and Julia~A. Noble.
\newblock Hierarchical class incremental learning of anatomical structures in
  fetal echocardiography videos.
\newblock \emph{IEEE Journal of Biomedical and Health Informatics}, 24\penalty0
  (4):\penalty0 1046--1058, April 2020.
\newblock \doi{10.1109/JBHI.2020.2973372}.

\bibitem[Perperidis et~al.(2017)Perperidis, Cusack, White, McDicken,
  MacGillivray, and Anderson]{perperidis2017dynamic}
A.~Perperidis, D.~Cusack, A.~White, N.~McDicken, T.~MacGillivray, and
  T.~Anderson.
\newblock Dynamic enhancement of b-mode cardiac ultrasound image sequences.
\newblock \emph{Ultrasound in Medicine \& Biology}, 43\penalty0 (7):\penalty0
  1533--1548, jul 2017.
\newblock \doi{10.1016/j.ultrasmedbio.2017.03.006}.

\bibitem[Pu et~al.(2021)Pu, Zhu, Li, and Li]{hybridclass}
Bin Pu, Ningbo Zhu, Kenli Li, and Shengli Li.
\newblock Fetal cardiac cycle detection in multi-resource echocardiograms using
  hybrid classification framework.
\newblock \emph{Future Generation Computer Systems}, 115:\penalty0 825--836, 02
  2021.
\newblock \doi{10.1016/j.future.2020.09.014}.

\bibitem[Qiao et~al.(2022)Qiao, Pang, Luo, Pan, Chen, and Lv]{qiao2022flds}
Sibo Qiao, Shanchen Pang, Gang Luo, Silin Pan, Taotao Chen, and Zhihan Lv.
\newblock Flds: An intelligent feature learning detection system for
  visualizing medical images supporting fetal four-chamber views.
\newblock \emph{IEEE Journal of Biomedical and Health Informatics}, 26\penalty0
  (10):\penalty0 4814--4825, October 2022.
\newblock \doi{10.1109/JBHI.2021.3091579}.
\newblock Epub 2022 Oct 4; PMID: 34156957.

\bibitem[Ronneberger et~al.(2015)Ronneberger, Fischer, and
  Brox]{ronneberger2015u}
Olaf Ronneberger, Philipp Fischer, and Thomas Brox.
\newblock U-net: Convolutional networks for biomedical image segmentation.
\newblock In \emph{Medical image computing and computer-assisted
  intervention--MICCAI 2015: 18th international conference, Munich, Germany,
  October 5-9, 2015, proceedings, part III 18}, pages 234--241. Springer, 2015.

\bibitem[Saha et~al.(2025)Saha, Mishra, Hernandez-Cruz, Patey, Papageorghiou,
  Asano, and Noble]{saha2025selfsupervisednormalitylearningdivergence}
Pramit Saha, Divyanshu Mishra, Netzahualcoyotl Hernandez-Cruz, Olga Patey, Aris
  Papageorghiou, Yuki~M. Asano, and J.~Alison Noble.
\newblock Self-supervised normality learning and divergence vector-guided model
  merging for zero-shot congenital heart disease detection in fetal ultrasound
  videos, 2025.
\newblock URL \url{https://arxiv.org/abs/2503.07799}.

\bibitem[Sharma et~al.(2021)Sharma, Drukker, Chatelain, Droste, Papageorghiou,
  and Noble]{SHARMA2021101973}
Harshita Sharma, Lior Drukker, Pierre Chatelain, Richard Droste, Aris~T.
  Papageorghiou, and J.~Alison Noble.
\newblock Knowledge representation and learning of operator clinical workflow
  from full-length routine fetal ultrasound scan videos.
\newblock \emph{Medical Image Analysis}, 69:\penalty0 101973, 2021.
\newblock ISSN 1361-8415.
\newblock \doi{https://doi.org/10.1016/j.media.2021.101973}.
\newblock URL
  \url{https://www.sciencedirect.com/science/article/pii/S1361841521000190}.

\bibitem[Ullah et~al.(2024)Ullah, Yan, and Fuxin]{ullah2024cvae_sm}
Amin Ullah, Taiqing Yan, and Li~Fuxin.
\newblock {CVAE-SM: A Conditional Variational Autoencoder with Style Modulation
  for Efficient Uncertainty Quantification}.
\newblock In \emph{Proceedings of the 2024 IEEE International Conference on
  Robotics and Automation (ICRA)}, pages 10786--10792, 2024.
\newblock \doi{10.1109/ICRA57147.2024.10611160}.

\bibitem[Upadhyay et~al.(2023)Upadhyay, Karthik, Mancini, and
  Akata]{upadhyay2023probvlmprobabilisticadapterfrozen}
Uddeshya Upadhyay, Shyamgopal Karthik, Massimiliano Mancini, and Zeynep Akata.
\newblock Probvlm: Probabilistic adapter for frozen vision-language models,
  2023.
\newblock URL \url{https://arxiv.org/abs/2307.00398}.

\bibitem[van~der Linde et~al.(2011)van~der Linde, Konings, Slager, Witsenburg,
  Helbing, Takkenberg, and Roos‐Hesselink]{vanderlinde2011birth}
D.~van~der Linde, E.~E.~E. Konings, M.~A. Slager, M.~Witsenburg, W.~A. Helbing,
  J.~J.~M. Takkenberg, and J.~W. Roos‐Hesselink.
\newblock Birth prevalence of congenital heart disease worldwide: a systematic
  review and meta-analysis.
\newblock \emph{Journal of the American College of Cardiology}, 58\penalty0
  (21):\penalty0 2241--2247, November 2011.
\newblock \doi{10.1016/j.jacc.2011.08.025}.
\newblock PMID: 22078432.

\bibitem[Wang et~al.(2021)Wang, Li, Zhou, and O.~Ogunbona]{opticalflow}
Zhen Wang, Guangxu Li, Jingjie Zhou, and Philip O.~Ogunbona.
\newblock Optical flow networks for heartbeat estimation in 4d ultrasound
  images.
\newblock In \emph{Proceedings of the 2021 7th International Conference on
  Computing and Artificial Intelligence}, ICCAI '21, page 127–131, New York,
  NY, USA, 2021. Association for Computing Machinery.
\newblock ISBN 9781450389501.
\newblock \doi{10.1145/3467707.3467725}.
\newblock URL \url{https://doi.org/10.1145/3467707.3467725}.

\bibitem[Wu et~al.(2024)Wu, Li, Chen, Ji, Lin, Gao, Kuang, Zhao, and
  Wu]{wu2024semanticalignmentmultimodallarge}
Tao Wu, Mengze Li, Jingyuan Chen, Wei Ji, Wang Lin, Jinyang Gao, Kun Kuang,
  Zhou Zhao, and Fei Wu.
\newblock Semantic alignment for multimodal large language models, 2024.
\newblock URL \url{https://arxiv.org/abs/2408.12867}.

\bibitem[Yang et~al.(2022)Yang, Zhang, Zhu, Wang, An, Gu, Liu, Han, He, and
  Zhu]{tenfetalseg}
Tingyang Yang, Ye~Zhang, Mengxiao Zhu, Yan Wang, Shan An, Xiaoyan Gu, Xiaowei
  Liu, Jiancheng Han, Yihua He, and Haogang Zhu.
\newblock Segmentation of ten fetal heart components with coarse‐to‐fine
  cascading and dynamic feature powering.
\newblock \emph{IET Image Processing}, 16:\penalty0 n/a--n/a, 07 2022.
\newblock \doi{10.1049/ipr2.12597}.

\bibitem[Yun(2011)]{Yun2011}
Sunhee~W. Yun.
\newblock Congenital heart disease in the newborn requiring early intervention.
\newblock \emph{Korean Journal of Pediatrics}, 54\penalty0 (5):\penalty0
  183--191, May 2011.
\newblock \doi{10.3345/kjp.2011.54.5.183}.
\newblock Epub 2011 May 31.

\bibitem[Zhang and Ré(2022)]{zhang2022contrastiveadaptersfoundationmodel}
Michael Zhang and Christopher Ré.
\newblock Contrastive adapters for foundation model group robustness, 2022.
\newblock URL \url{https://arxiv.org/abs/2207.07180}.

\end{thebibliography}

\newpage
\appendix
\section*{Supplementary Material}  
In the Supplementary Material, we first present in Section A detailed descriptions of private CHD dataset and of the EchoNet-Dynamic dataset, including how we converted its ejection-fraction regression labels into a classification task. Section B defines all evaluation metrics (F1, AUC, ECE, AECE). Section C covers implementation details such as frame sampling, optimizer settings, hyperparameters, and prompt templates. Finally, Section D reports the complete ablation results. Our code will be released publicly upon publication.

\section{Datasets}
\label{app:datasets}

\noindent\textbf{Private Dataset.}  
As no public benchmark for fetal CHD detection exists, a private dataset of 477 echocardiographic videos was curated. Each video, showing the four-chamber (4CH) view of the fetal heart, was acquired at a local hospital and annotated by a senior Maternal–Fetal Medicine consultant. Videos were labeled as either Normal (198 cases) or as one of 19 distinct congenital heart disease (CHD) classes. Among these, the most frequently represented defects are ventricular septal defect (VSD) (43 cases), atrioventricular septal defect (AVSD) (33), arrhythmia (33), and cardiomegaly (31). Model performance was assessed via 5-fold cross-validation with class-stratified splits.

\noindent\textbf{EchoNet-Dynamic Dataset.}  
The EchoNet-Dynamic dataset \cite{ouyang2020video}, comprising 10,030 echocardiographic videos annotated with estimated ejection fraction (EF), was employed for further evaluation. Following the PanEcho protocol \cite{panecho}, EF values were binned into three systolic dysfunction categories: ``Moderately/Severely Decreased'' (EF $<40\%$), ``Mildly Decreased'' ($40\%\le \mathrm{EF}\le 54\%$), and ``Normal/Hyperdynamic'' (EF $>54\%$).

\section{Metrics}
\label{app:eval}

Discrimination was assessed using the macro F1 score, which balances precision and recall, and the area under the receiver operating characteristic curve (AUC), which quantifies the ability to rank positive instances above negative ones across all thresholds. Calibration was measured by the expected calibration error (ECE), defined as the average absolute difference between predicted confidence and empirical accuracy over fixed bins, and by the adaptive expected calibration error (AECE), which employs quantile-based bins to prevent underpopulated intervals.
\begin{description}[leftmargin=1.5cm,labelsep=0.5cm]
  \item[F1 Score] The harmonic mean of precision and recall,
    \begin{equation}\label{eq:f1}
      \mathrm{F1} = 2 \cdot \frac{\mathrm{precision}\,\cdot\,\mathrm{recall}}
                                       {\mathrm{precision} + \mathrm{recall}}
    \end{equation}
    which balances false positives and false negatives.

  \item[Macro F1 Score] The unweighted average of per‐class F1 scores,
    \begin{equation}\label{eq:macro_f1}
      \mathrm{F1}_{\mathrm{macro}}
        = \frac{1}{C}\sum_{c=1}^{C}\mathrm{F1}_c
    \end{equation}
    giving equal weight to each class in multiclass settings.

  \item[Area under the ROC Curve (AUC)] The probability that a randomly chosen positive instance is ranked higher than a randomly chosen negative one. Values range from 0.5 (random) to 1.0 (perfect). For multiclass tasks, we use the macro‐average of one‐vs‐rest AUCs.

  \item[Expected Calibration Error (ECE)] \cite{guo2017calibration}
    \begin{equation}\label{eq:ece}
      \mathrm{ECE} = \sum_{m=1}^{M}\frac{|B_{m}|}{n}
                             \bigl|\mathrm{acc}(B_{m}) - \mathrm{conf}(B_{m})\bigr|
    \end{equation}
    where \(B_m\) is the set of predictions in bin \(m\), \(n\) is the total number of samples, and \(\mathrm{acc}(B_m)\) and \(\mathrm{conf}(B_m)\) are its accuracy and average confidence.  
    In our experiments we set \(M=15\) uniform‐width bins.

  \item[Adaptive Expected Calibration Error (AECE)] \cite{nixon2019measuring}  
    Computed with quantile‐based bins so each bin has roughly the same number of samples. 
\end{description}

\section{Implementation Details}
\label{app:implementation}

\noindent\textbf{Implementation Details.} Following \cite{panecho,fetalCLIP,ehad456}, \(L=16\) consecutive frames were sampled from each video. A batch size of 16 was used, the Adam optimizer (initial learning rate \(1\times10^{-3}\)) was employed, and a ReduceLROnPlateau scheduler (decay factor 0.1, patience 5) was applied. On the private CHD dataset, training was carried out for up to 40 epochs with early stopping based on validation macro F1, for EchoNet-Dynamic, the same hyperparameters were retained, but training was restricted to 20 epochs. All experiments were conducted on an NVIDIA RTX 5000 Ada Generation GPU. For uncertainty quantification, \(\beta\) was set to 0.2 in Equation~\ref{eq:total_loss_uncertainty}.  

\smallskip\noindent\textbf{Binary CHD detection prompts.}  For binary CHD detection we used, ``Is the fetal heart normal in this 4CH ultrasound view?'' and ``Can an abnormality be detected in the fetal heart 4CH?''
\noindent For prompt randomization, we sample one prompt per class from six variants:
\begin{description}
  \item[Normal:] 
    “Is the fetal heart normal in this 4CH ultrasound view?”;
    “Does this 4CH image show a normal heart condition?”;
    “Is everything normal in this fetal heart 4CH scan?”;
    “Does this ultrasound suggest a healthy fetal heart?”;
    “Does this 4CH scan reflect a structurally normal heart?”;
    “normal”
  \item[Abnormal:] 
    “Is abnormality visible in this fetal heart 4CH scan?”;
    “Does this image show abnormal fetal heart?”;
    “Can abnormality be detected in this fetal heart 4CH?”;
    “Does this 4CH scan suggest the presence of abnormality?”;
    “Does this ultrasound reveal abnormality in the fetal heart?”;
    “abnormal”
\end{description}

\noindent\textbf{Multiclass prompts.} Each defect class is associated with one fixed, class-specific description:
\begin{description}[leftmargin=1.5cm,labelsep=0.5cm]
  \item[VSD] 
    A defect in the interventricular septum permits blood to flow from the left to the right ventricle, causing abnormal left-to-right shunting and right-heart volume overload.
  \item[AVSD] 
    This anomaly features incomplete fusion of the atrioventricular septum and malformed AV valves, resulting in a common AV junction and mixed atrial-ventricular flow.
  \item[Arrhythmia] 
    Rhythmic irregularities demonstrate arrhythmia, from isolated premature beats to sustained tachy- or bradyarrhythmic patterns.
  \item[Cardiomegaly] 
    An enlarged cardiac silhouette occupying over half the thoracic area indicates cardiomegaly, often due to volume overload, cardiomyopathy, or high-output states.
\end{description}

\noindent\textbf{EchoNet-Dynamic prompts.} We use three fixed descriptions that align with the ejection‐fraction categories:
\begin{itemize}[nosep]
  \item ``Normal/Hyperdynamic''
  \item ``Mildly Decreased''
  \item ``Moderately/Severely Decreased''
\end{itemize}

\section{Ablation}
\label{app:ablations}
\begin{table}[ht]
  \centering
  \scriptsize
  \renewcommand{\arraystretch}{1.2}

  \begin{minipage}{0.40\textwidth}
    \centering
    \setlength{\tabcolsep}{8pt}
    \begin{tabularx}{\textwidth}{@{}l>{\hspace{10pt}}S
      S[table-format=2.2]
      S[table-format=2.2]@{}}
      \toprule
      \textbf{Hyperparameter}
        & {\textbf{F1}($\uparrow$)}
        & {\textbf{AUC}($\uparrow$)} \\
      \midrule
      $m=0.5$         & \multicolumn{1}{S[table-format=2.2]}{\bfseries 84.69} & 86.14 \\
      $m=1.0$         & 84.45 & \bfseries 86.88 \\
    \midrule
      $\alpha=1$     & 84.20 & 86.89 \\
      $\alpha=0.5$ & \multicolumn{1}{S[table-format=2.2]}{\bfseries 84.69} & \multicolumn{1}{S[table-format=2.2]}{\bfseries 87.59} \\
      $\alpha=0$     & 84.67 & 86.11 \\
    \midrule
      random prompts  & 84.24 & 86.96 \\
      \bottomrule
    \end{tabularx}
        \vspace{3mm}
    \caption*{(a) $m$, $\alpha$, and prompt randomization.}
  \end{minipage}
  \hfill
  \begin{minipage}{0.58\textwidth}
    \centering
    \setlength{\tabcolsep}{20pt}
    \begin{tabularx}{\textwidth}{@{}l
      S[table-format=1]
      S[table-format=1]@{}}
      \toprule
      \textbf{Hyperparameter}
        & {\textbf{F1}($\uparrow$)}
        & {\textbf{AUC}($\uparrow$)} \\
      \midrule
      4 neighbors, mean pooling & 83.42 & 86.55 \\
      16 neighbors, mean pooling & 83.71 & 86.92 \\
      10 neighbors, mean pooling & 84.42 & \bfseries 89.02 \\
      10 neighbors, max pooling & \bfseries 85.01 & 87.49 \\
      3 passes through GNN       & 84.31 & 87.71 \\
      graph fusion by summation      & 84.58 & 88.46 \\
      \bottomrule
    \end{tabularx}
        \vspace{6.6mm}
    \caption*{(b)Ablation on GNN components.}
    \label{tab:gnn_ablations}
  \end{minipage}
  \vspace{4mm}

  \begin{minipage}[ht]{0.40\textwidth}
    \centering
    \setlength{\tabcolsep}{10pt}
    \begin{tabularx}{\textwidth}{@{}l
      S[table-format=1]
      S[table-format=1]@{}}
      \toprule
      \textbf{Hyperparameter}
        & {\textbf{F1}($\uparrow$)}
        & {\textbf{AUC}($\uparrow$)} \\
      \midrule
      mLSTM           & 84.98 & 87.70 \\
      sLSTM           & \bfseries 85.43 & \bfseries 88.33 \\
      mid fusion      & 84.82 & 86.31 \\
      late fusion     & 85.15 & 88.06 \\
      2-block         & 84.46 & 85.69 \\
    7-block & 85.40 & 88.31 \\
      12-block        & 85.35 & 86.10 \\
      \bottomrule
    \end{tabularx}
        \vspace{3mm}
    \caption*{(c) Ablation on xLSTM components.}
    \label{tab:xlstm_ablations}
  \end{minipage}
  \hfill
\begin{minipage}[ht]{0.58\textwidth}
  \centering
  \setlength{\tabcolsep}{6pt}
  \sisetup{
    table-format=1,
    table-align-text-post = false,
  }
  \renewcommand{\arraystretch}{1.2}
  \begin{tabularx}{\textwidth}{@{}l
    S[table-format=1] S[table-format=1]
    S[table-format=1] S[table-format=1]@{}}
    \toprule
    \textbf{Temporal Extractor}
      & \multicolumn{2}{c}{\textbf{Private dataset}}
      & \multicolumn{2}{c}{\textbf{EchoNet-Dynamic}} \\
    \cmidrule(lr){2-3} \cmidrule(lr){4-5}
    & {F1} & {AUC} & {F1} & {AUC} \\
    \midrule
    1D-CNN      & 81.56    & 82.44    & 56.68 & 81.38 \\
    TCN         & 81.83   & 85.44   & 55.98 & 80.51 \\
    GNN         & \bfseries 84.84    & \bfseries 87.69   & \bfseries 58.62 & \bfseries 82.11 \\
    xLSTM       & 80.32    & 84.64   & 54.01 & 79.51 \\
    \bottomrule
  \end{tabularx}
  \vspace{3.6mm}
  \caption*{(d) Ablation on Temporal Extractors within TPA.}
 \label{tab:echonet_ablations}
  
\end{minipage}
\vspace{3mm}
\caption{
Ablation study of TPA performance: (a) Loss hyperparameters: $m$, $\alpha$ in Equation \ref{eq:total_loss}, and prompt randomization; 
(b) GNN design choices: number of neighbors, pooling method and fusion strategy; 
(c) xLSTM variants: gating type (mLSTM vs sLSTM), fusion stage and network depth; 
(d) Comparison of different temporal extractors within TPA on the private and EchoNet‐Dynamic datasets. Best results are highlighted in \textbf{bold}.
}
  \label{tab:tpa_grid_ablation_clean}
\end{table}

\begin{table}[ht]
  \centering
  \scriptsize
  \setlength{\tabcolsep}{6pt}
  \renewcommand{\arraystretch}{1.2}
  \sisetup{
    table-format=2.4(2),
    table-align-text-post = false,
  }
  \begin{tabularx}{\textwidth}{@{}X
    S[table-format=2.2(4)] S[table-format=2.2(4)]
    S[table-format=2.2(4)] S[table-format=2.2(4)]
    S[table-format=2.2(4)] S[table-format=2.2(4)]@{}}
    \toprule
    \textbf{Class}
      & \multicolumn{2}{c}{\textbf{TCN}}
      & \multicolumn{2}{c}{\textbf{GNN}}
      & \multicolumn{2}{c}{\textbf{xLSTM}} \\
    \cmidrule(lr){2-3} \cmidrule(lr){4-5} \cmidrule(lr){6-7}
    & {F1 $(\uparrow)$} & {AUC $(\uparrow)$} & {F1 $(\uparrow)$} & {AUC $(\uparrow)$} & {F1 $(\uparrow)$} & {AUC $(\uparrow)$} \\
    \midrule 
    \multicolumn{7}{@{}l}{\bfseries Defects = 2} \\
    Macro      & $\mathbf{78.68 \pm 8.20}$ & 71.09(11.35) & 76.46(9.92) & 72.76(13.72) & 76.82(8.73) & 72.55(14.69) \\
    Normal     & $\mathbf{88.69 \pm 5.24}$ & 76.36(9.69)  & 88.09(6.55) & 78.25(11.76) & 88.34(6.54) & 79.22(13.07) \\
    VSD        & $\mathbf{53.96 \pm 1.13}$ & 62.93(1.63)  & 48.78(17.42) & 65.06(20.30) & 51.69(1.54) & 65.92(1.96) \\
    AVSD       & $\mathbf{59.56 \pm 2.06}$ & 73.98(14.78) & 59.56(18.33) & 74.96(14.79) & 57.80(1.21) & 67.95(1.55) \\
    \midrule
    \multicolumn{7}{@{}l}{\bfseries Defects = 3} \\
    Macro      & $\mathbf{72.91 \pm 3.75}$  & 70.07(8.94)  & 71.80(4.60)  & 69.53(8.13)  & 71.27(6.03)  & 69.37(9.94) \\
    Normal     & 84.62(2.47)  & 83.79(9.54)  & 85.08(2.92)  & 79.50(7.11)  & $\mathbf{85.22 \pm 4.81}$  & 77.44(8.41) \\
    VSD        & $\mathbf{56.86 \pm 15.34}$ & 70.01(19.50) & 51.14(10.78) & 69.97(20.50) & 54.50(1.33)  & 73.24(2.02) \\
    AVSD       & $\mathbf{59.26 \pm 16.84}$ & 69.83(17.06) & 56.03(8.92)  & 68.56(15.47)        & 56.16(1.21)  & 67.95(1.15) \\
    Arrhythmia & 43.99(11.13) & 56.64(10.69) & 44.17(15.02) & 60.09(12.77) & $\mathbf{45.78 \pm 1.34}$  & 58.85(1.32) \\
    \midrule
    \multicolumn{7}{@{}l}{\bfseries Defects = 4} \\
    Macro        & $\mathbf{67.63 \pm 5.33}$  & 66.20(5.46)  & 67.49(5.18)  & 66.77(2.71)  & 66.90(4.64)  & 51.61(10.63) \\
    Normal       & 84.72(2.77)  & 86.61(5.43)  & $\mathbf{86.61 \pm 3.23}$  & 86.48(6.58)  & 84.49(4.04)  & 86.64(4.99) \\
    VSD          & $\mathbf{46.70 \pm 11.88}$ & 57.70(17.77) & 46.61(11.59) & 60.33(14.20) & 43.27(8.60)  & 58.23(19.00) \\
    AVSD         & $\mathbf{55.98 \pm 17.38}$ & 68.19(16.30) & 51.94(7.93)  & 66.27(12.18) & 46.32(18.12) & 62.47(15.51) \\
    Arrhythmia   & 31.14(9.88)  & 42.80(12.87) & 34.21(7.36)  & 47.24(12.22) & $\mathbf{37.53 \pm 9.58}$  & 49.27(9.44) \\
    Cardiomegaly & 55.68(17.09) & 75.72(16.74) & 54.77(10.63) & 73.52(9.09)  & $\mathbf{58.68 \pm 14.38}$ & 69.59(1.38) \\
    \bottomrule
  \end{tabularx}
  \vspace{3mm}
    \caption{
    Ablation of temporal feature extractors in TPA on the private CHD dataset for multiclass classification with 2, 3, and 4 defects plus the normal class, reporting macro‐F1 and AUC across 5 folds. Best results are highlighted in \textbf{bold}.
    }
  \label{tab:tpa_f1_auc_multiclass_grouped_tpa}
\end{table}

\end{document}